\documentclass[twocolumn]{svjour3}          

\smartqed  
\usepackage[T1]{fontenc}
\usepackage[utf8]{inputenc}
\usepackage{newtxtext,newtxmath}
\usepackage{graphicx}
\usepackage{epstopdf}
\usepackage{epsfig,graphics}
\usepackage{latexsym}
\usepackage{multirow}
\usepackage{booktabs}
\usepackage{cite}
\usepackage{citeref}
\usepackage{subfigure}
\usepackage[misc]{ifsym}
\usepackage[export]{adjustbox}
\usepackage[T1]{fontenc}
\usepackage{hyperref}
\usepackage{mathtools,xparse}
\usepackage{pgfplots}
\usepackage{tikz}
\usepackage{tkz-euclide}
\usetkzobj{all}
\usetikzlibrary{positioning}
\usetikzlibrary{shapes,arrows}
\usetikzlibrary{shadows, positioning,arrows.meta,shapes.geometric}
\usepackage[flushleft]{threeparttable}
\usepackage{floatrow}
\usepackage{wrapfig}
\usepackage{algorithm}
\usepackage[noend]{algpseudocode}
\usepackage{etex}

\begin{document}

\title{Regularized HessELM and Inclined Entropy Measurement for Congestive Heart Failure Prediction}




\author{Apdullah Yay{\i}k \and Yakup Kutlu \and G\"{o}khan Altan}



\institute{A. Yay{\i}k (\Letter)  \at Military Academy \\National Defence University, Ankara, Turkey\\  email: ayayik@kho.edu.tr\\
Y. Kutlu \at Department of Computer Engineering,\\\.{I}skenderun Technical University, Hatay, Turkey\\
G. Altan \at Department of Computer Engineering,\\\.{I}skenderun Technical University, Hatay, Turkey
}


\maketitle
\begin{abstract}
Our study concerns with automated predicting of congestive heart failure (CHF) through the analysis of electrocardiography (ECG) signals. A novel machine learning approach, regularized hessenberg decomposition based extreme learning machine (R-HessELM), and feature models; squared, circled, inclined and grid entropy measurement were introduced and used for prediction of CHF. This study proved that inclined entropy measurements features well represent characteristics of ECG signals and together with R-HessELM approach overall accuracy of 98.49\% was achieved.

\keywords{Congestive heart failure \and Entropy measurements \and Extreme learning machine \and Hessenberg decomposition \and Regularization}
\end{abstract}

\section{Introduction}
\label{intro}
Congestive heart failure (CHF) is a serious medical condition (not a disease) that inhibits the heart performing the circulatory activities throughout the body, since the heart can not pump sufficient blood that tissues need. Its symptoms are generally breathless, ankle swelling and fatigue. It is associated with significant decrease in quality of life and high degrees of debility, morbidity and mortality. According to British Society for Heart Failure report on March 2016, in United Kingdom approximately 900.000 people are exposed to CHF, 5\% of emergency admissions are related with  CHF and its treatment processes  consume 2\% of all national health service expenditure \cite{ref1}. Since it is epidemic not only across Europe also worldwide, the need of early diagnosis (prediction its existence) is a very important issue. Unfortunately, it could not be assessed with ease using clinical techniques. European Society of Cardiology guidelines in 2016 recommend the following procedure to diagnose CHF. Clinical history of the patient,  physical examination, characterized symptoms also electrocardiography (ECG) record while resting are analyzed. If at least one of the components is not normal,  plasma Natriuretic Peptides is to be examined to decide whether echocardiography is needed \cite{ref2}. In order to provide rapid and reliable  diagnosis,  automatic prediction based on data mining or machine learning techniques using only  ECG data is a vital research area. 

For the last decade, several signal processing and machine learning approaches were utilized on ECG signal to predict CHF. Researchers heavily focused on revealing robust features from power spectrum of ECG signal. In particular, statistical values of  discrete wavelet transform \cite{ref3, ref4} wavelet decomposition \cite{ref5} and continuous wavelet transform \cite{ref6} were extracted. Whereas, few studies focused on time-series properties of ECG signal, such as Poincare \cite{ref7, ref8}, R$-$R intervals \cite{ref9,ref10}, second-order difference plot (SODP) \cite{ref11, ref12}. \.{I}\c{s}ler and Kuntalp analyzed heart-rate variability (HRV) with many feature extracting techniques not only based on power spectrum also time-frequency approaches \cite{ref13}. Measure of complex correlation to quantify temporal variability in the Poincare plot was introduced by Karmakar et al. \cite{ref14}. They reported efficiency of central tendency measure of the R$-$R interval and radial distance of Teager energy scatter plot to predict CHF. Thuraisingham \cite{ref15} analyzed SODP of R$-$R intervals. In addition, he introduced a classification system that employs statistical procedure. Cohen et al.\cite{ref12} analyzed SODPs and central tendency measures. Zheng et al. proposed least square support vector machines (LS-SVM) model with heart sound and cardiac reverse features that predict CHF. Additionally, they showed that it outperforms neural network and hidden-markow model \cite{ref16}. Altan et al proposed an early diagnosis model for CHF by applying Hilbert-Huang Transform to the ECG signal. They extracted high-order statistical features form several frequency modulations \cite{ref17}.

Our study concerns with prediction of CHF with a reliable intelligence system. Novel approaches for distributing peaks named squared entropy measurement (SEM), inclined entropy measurement (IEM) and grid entropy measurement (GEM) for extracting several scattered features were introduced. In back-propagation (BP) algorithm, minimizing loss function is controlled using a different data, namely validation set, excluded from training set to avoid over-fitting, in other words to make it gain generalization capability. However, extreme learning machine (ELM) cannot gain a generalization capability using necessarily the same approach in BP. Recently, Cao et al.  proposed that ELM with singular value decomposition (SVD) method can gain generalization ability with considering performance of pseudoinversing in leave-one out (LOO) model when computing loss function $-$mean square error (MSE) \cite{ref18}. Therefore, they put regularization term to the denominator of classical MSE, in order to optimally select regularization term by observing effect of pseudoinversing. In this study this approach is extended to recently introduced hessenberg decomposition based ELM in \cite{ref18}.

In the following section, the data acquisition, preprocessing steps, methods of feature extraction, classifiers and performance measures are presented. In Sect. \ref{resultsandconclusions}, results are given and analyzed. Discussions about the results are in Sect. \ref{discussion}.

\section{Materials and Methods} 
\label{sec:Materials and Methods}
\begin{figure}
    \centering
    \includegraphics[width=1\textwidth]{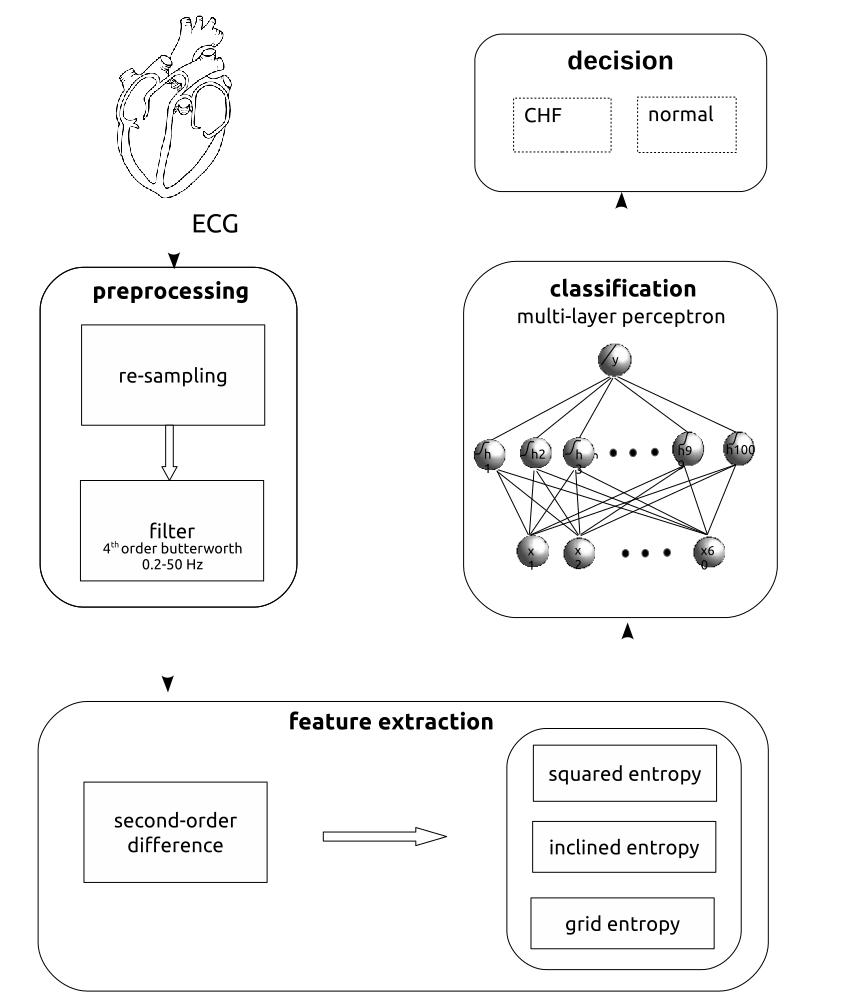}
    \caption{The proposed system for the predicting CHF}
    \label{figure1}
\end{figure}

The system architecture and data processing are illustrated in Figure \ref{figure1}.

\subsection{Data Acquisition}
In this study CHF Database and Normal Sinus Rhythm Database which are freely available on Physionet web site \cite{ref20} were used.  The CHF database consists of 24 hour ECG signals with 250 Hz sampling  frequency from 15 patients (aged between 22 and 71) exposed to CHF and  Normal sinus rhythm database consists of ECG signals with 128 Hz sampling frequency (resampled to 256 Hz with cubic splines) from 18 healthy people (aged between 20 and 50). ECG recordings had power line interference and baseline wander effects because of respiration. Baseline wander and Power line interference consist of low frequency components and high frequency components, respectively. ECG recordings were filtered with two median filters to eliminate the baseline wander and with a notch filter to eliminate power-line frequency \cite{ref21}. 

\subsection{Second-Order Difference Plot (SODP)}
Second-order difference plot (SODP) was originated from Poincare plot \cite{ref8}. It scatters consecutive difference values with second degree over the Cartesian coordinate system. For $\mathbf{X}=[\mathbf{x}_1, \mathbf{x}_2, \dots, \mathbf{x}_n]\epsilon \Re^n$ is data matrix where $x_i$ vectors are rows and n is the number of attributes. SODP values for $i^{th}$ row is calculated as follows,

\begin{equation}
\label{sodp}
    \begin{split}
        a &=x_{i+1}-x_i \\
        b &=x_{i+2}-x_{i+1} 
    \end{split}
\end{equation}

Plotting a against b in (\ref{sodp}) gives SODP. SODP provides observing the statistical situation of consecutive differences in time series data. In this study, four types of SODP whose peak-to-peak points were distributed geometrically distinct from each other were investigated.

\begin{figure}
    \centering
    \includegraphics[width=1\textwidth]{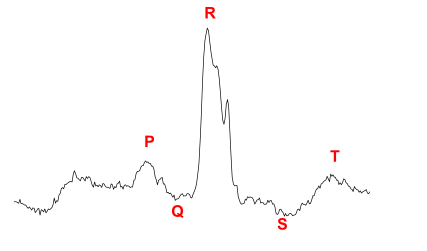}
    \caption{Peak representation of 1 s ECG data used in this study}
    \label{figure2}
\end{figure}

In order to predict CHF accurately, features that are related to characteristic structure of ECG recordings needs to be extracted. SODP distributes ECG data peaks in Figure \ref{figure2} into different regions of Cartesian coordinate system according to amplitude variability. While P and T peaks refer to little rate amplitude increase and decrease, Q and S peaks refer to little rate decrease and increase, R peaks refer to high rate amplitude increase and decrease on ECG signal. All peaks increase and decrease regularly, except for P and T peaks. In SODP; while P, T, Q and S peaks fall near and above $+$x axis and near and below $-$x axes, R peaks fall above and away from $+$x axes and below and away from $-$x axes. Therefore, increasing and decreasing rate, regularity of peak fluctuations and closeness to x axes are significant properties that help characterize ECG signal.

\begin{figure*}
  \centering
  \subfigure[]{%
    \label{fig:A}%
    \includegraphics[width=0.26\textwidth]{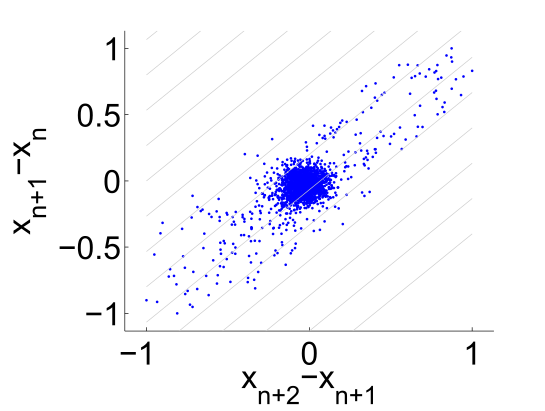}
            }%
  \subfigure[]{%
    \label{fig:B}%
   \includegraphics[width=0.26\textwidth]{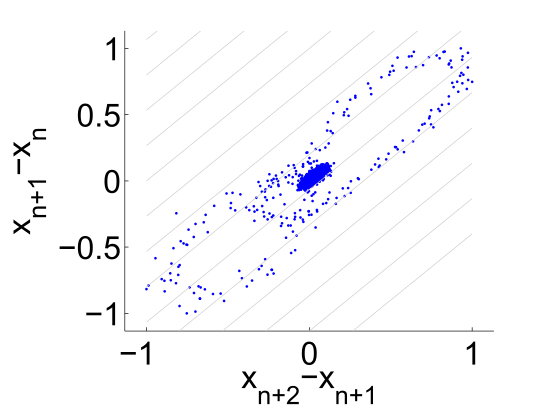}
            }%
  \subfigure[]{%
    \label{fig:C}%
    \includegraphics[width=0.27\textwidth]{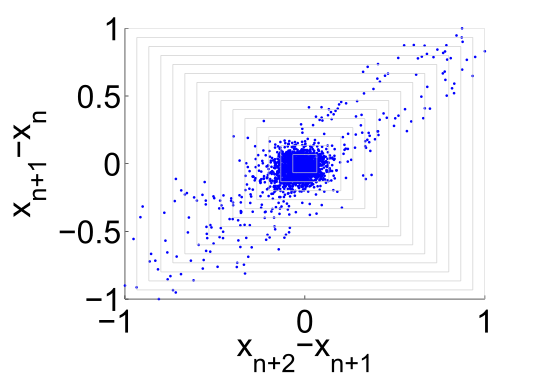}
            }%
  \subfigure[]{%
    \label{fig:D}%
    \includegraphics[width=0.25\textwidth]{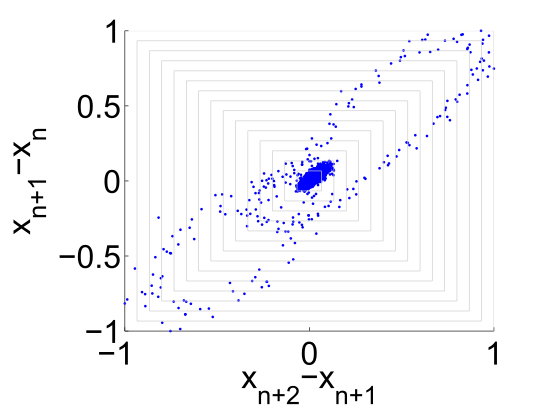}
           }\\
  \subfigure[]{%
    \label{fig:E}%
    \includegraphics[width=0.26\textwidth]{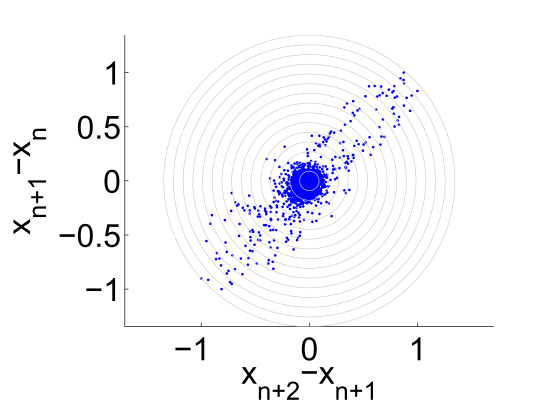}
            }%
  \subfigure[]{%
    \label{fig:F}%
    \includegraphics[width=0.26\textwidth]{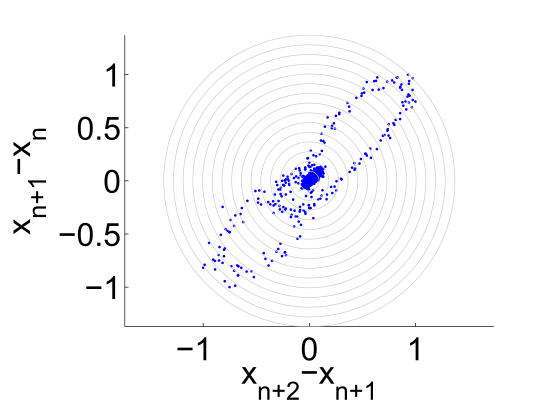}
            }%
  \subfigure[]{%
    \label{fig:G}%
    \includegraphics[width=0.26\textwidth]{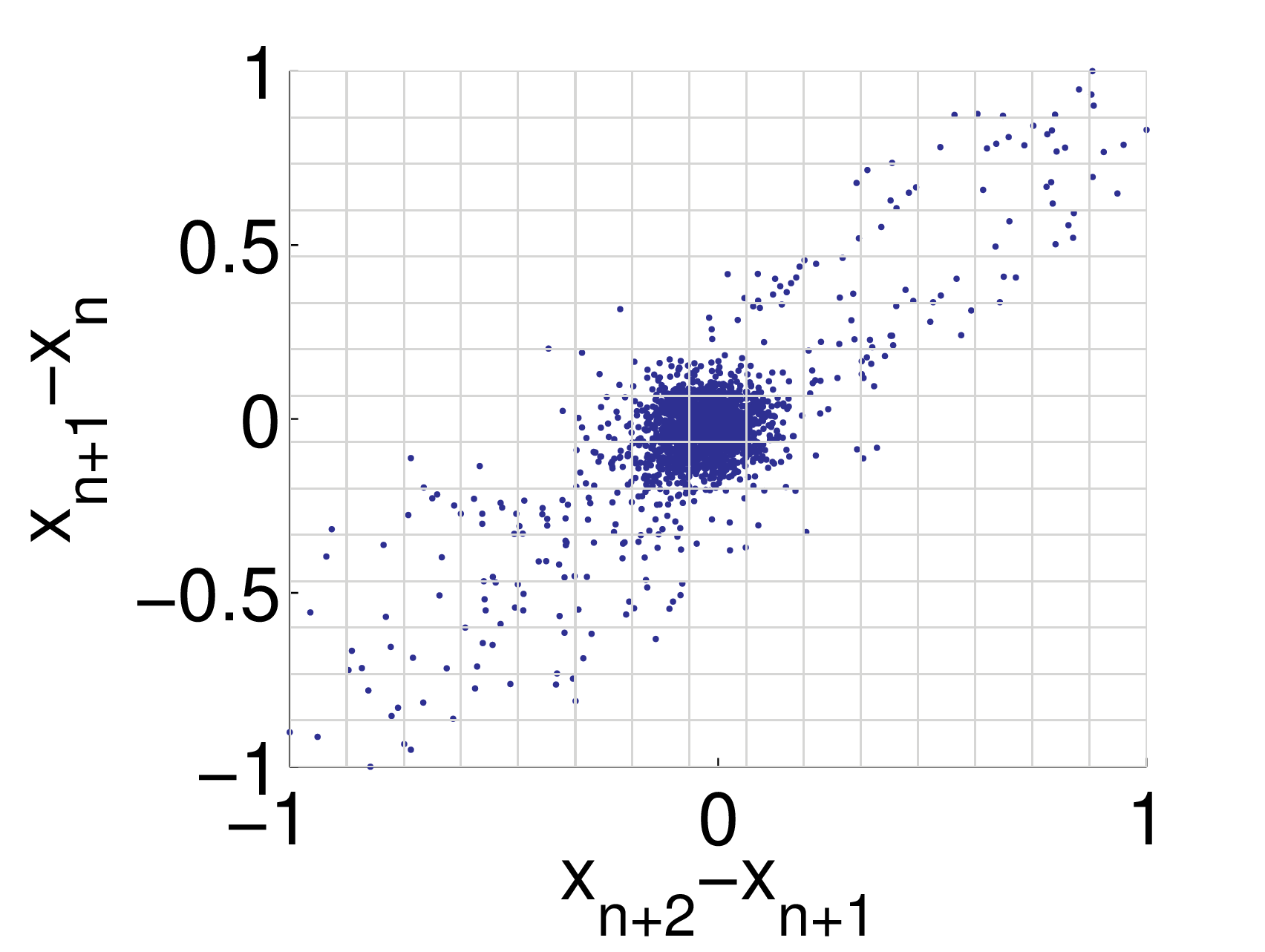}
            }%
  \subfigure[]{%
    \label{fig:H}%
    \includegraphics[width=0.26\textwidth]{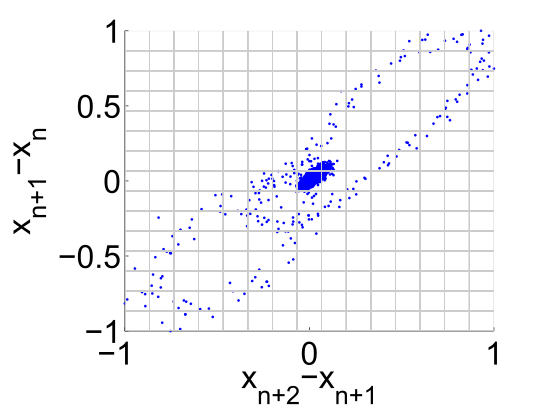}
            }
  \caption{SODP of ECG data (belonging to patients exposed to CHF) with (a) inclined, (c) squared, (e) circled (g) grid entropy measurements and SODP of  ECG data (belonging to healthy person) with (b) inclined, (d) squared, (f) circled, (h) grid entropy measurements}%
  \label{figure3}%
\end{figure*}

In this study; to extract discriminative information from SODP, cumulative number of data in circled, squared, inclined and grid regions were calculated. Novel feature models; circled entropy measurement (CEM) that calculates data fall over circled pieced regions,  squared entropy measurement (SEM) that calculates data fall over squared pieced regions, inclined entropy measurement (IEM) that calculates data fall over inclined pieced regions, grid entropy measurement (GEM) that calculates data fall over grid pieced regions in SODP were introduced. These feature models for CHF and Normal ECG data are indicated in Figure \ref{figure3}. SEM and CEM methods captures similar ratio of changes in ECG peaks with different accuracies. IEM captures wide-range of changes in ECG peaks. IEM method can detect number of regularly increasing and decreasing, only instantaneous increasing after decreasing and only instantaneous decreasing after increasing characteristic of ECG peaks. GEM captures local fluctuations of ECG data in Cartesian system. GEM detects number of fluctuations in ECG peaks according to axis distance information. MATLAB script for feature models is shared for practitioners \footnote{https://github.com/apdullahyayik/time-series-analysis}

\subsection{Normalization}
Prior to classification, features were linearly normalized using \ref{normalization} within the range $[-1, +1]$
\begin{equation}
    \label{normalization}
    x_{norm}=2\left(\frac{x-x_{min}}{x_{max}-x_{min}}\right)-1
\end{equation}
where $x_{max}$ and $x_{min}$ represent respectively the lowest and highest values of each feature. The normalization coefficients, which were extracted from the training data, were stored to normalize the test data as well.

\subsection{Extreme Learning Machine (ELM)}
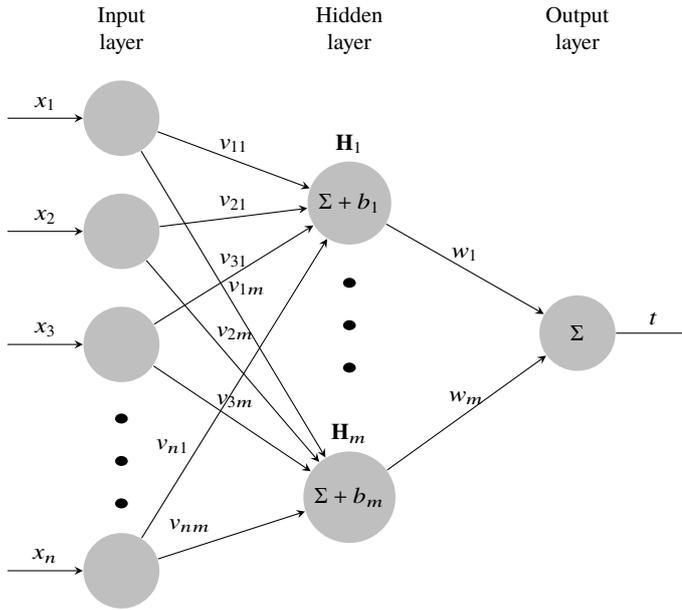
\begin{figure}
\centering
\tikzset{%
   neuron missing/.style={
    draw=none, 
    scale=4,
    text height=0.333cm,
    execute at begin node=\color{black}$\vdots$
  },
}

\begin{tikzpicture}[x=1.5cm, y=1.5cm, >=stealth]

\foreach \m/\l [count=\y] in {1,2,3}
{
 \node [circle,fill=gray!50,minimum size=1cm] (input-\m) at (0,2.5-\y) {};
}
\foreach \m/\l [count=\y] in {4}
{
 \node [circle,fill=gray!50,minimum size=1cm ] (input-\m) at (0,-2.5) {};
}
 
 \node [neuron missing]  at (0,-1.5) {};

\foreach \m [count=\y] in {1}
  \node [circle,fill=gray!50,minimum size=1cm ] (hidden-\m) at (2,0.75) {$\Sigma+b_1$};
  
\foreach \m [count=\y] in {2}
  \node [circle,fill=gray!50,minimum size=1cm ] (hidden-\m) at (2,-1.85) {$\Sigma+b_m$};
  
 \node [neuron missing]  at (2,-0.3) {};

\foreach \m [count=\y] in {1}
  \node [circle,fill=gray!50,minimum size=1cm ] (output-\m) at (4,0.6-\y) {$\Sigma$};
 

\foreach \l [count=\i] in {1,2,3,n}
  \draw [<-] (input-\i) -- ++(-1,0)
    node [above, midway] {$x_{\l}$};

\foreach \l [count=\i] in {1,m}
  \node [above] at (hidden-\i.north) {$\mathbf{H}_{\l}$};

\foreach \l [count=\i] in {1}
  \draw [->] (output-\i) -- ++(1,0)
    node [above, midway] {$t$};

  
    \draw [->] (input-1) -- (hidden-1)
    node[above, midway]{$v_{11}$};
    \draw [->] (input-1) -- (hidden-2)
    node at (1.1,0) {$v_{1m}$};
   
   \draw [->] (input-2) -- (hidden-1)
    node[above, midway]{$v_{21}$};
    \draw [->] (input-2) -- (hidden-2)
     node at (1,-0.4) {$v_{2m}$};
    
    \draw [->] (input-3) -- (hidden-1)
    node[above, midway]{$v_{31}$};
    \draw [->] (input-3) -- (hidden-2)
   node at (1,-1){$v_{3m}$};
    
    \draw [->] (input-4) -- (hidden-1)
    node at (0.45,-1.4) {$v_{n1}$};
    \draw [->] (input-4) -- (hidden-2)
    node at (0.6,-2.1) {$v_{nm}$};
    
    \draw [->] (hidden-1) -- (output-1)
     node[above, midway]{$w_1$};
     \draw [->] (hidden-2) -- (output-1)
     node[above, midway]{$w_m$};

\foreach \l [count=\x from 0] in {Input, Hidden, Output}
\node [align=center, above] at (\x*2,2) {\l \\ layer};

\end{tikzpicture}
\caption{Structure of ELM}
\label{figure4}
\end{figure}

Conventional ELM \cite{ref22} is a fully-connected single-hidden layer feed-forward neural network (SLFN) that has random number of nodes in hidden layer. Its structure is illustrated in Figure \ref{figure4}. In the input layer, weights and biases are assigned randomly, whereas in the output layer, weights are computed with non-iterative linear optimization technique based on generalized-inverse. Hidden layer with non-linear activation function makes non-linear input data linearly-separable. Let ${(x_{i},t_{i})}$ be a sample set, with n distinct samples, where $\mathbf{x}_i=[x_{i1}, x_{i2}, \dots, x_{in}]^T$  is the $i^{th}$ is $i^{th}$input sample and  is the $i^{th}$ desired output. With m hidden neurons, the output of $k^{th}$ hidden layer is given by (where k<m);

\begin{equation}
    \mathbf{H}_{ik}=\varphi(\sum_{i=1}^n x_{ik}{v}_{ik}+b_{k})
\label{elm1}
\end{equation}
 and $k^{th}$ desired output is given by;
 \begin{equation}
     \mathbf{t_{k}}= \sum_{j=1}^mH_{jk}{w}_j 
      \label{elm2}
\end{equation}

\begin{equation}
    \mathbf{H}w=t
    \label{elm3}
\end{equation}
where $\varphi(\cdot)$ is the activation function, $\mathbf{H}=[h_{i1} \dots h_{im}]$ is the output of hidden neurons, $\mathbf{v}=[v_{i1}, \dots, v_{in}]$ is the input layer weight matrix, $\mathbf{w}=[w_{1},\dots, w_{m}]^T$ is the output layer weight matrix, $\mathbf{b}=[b_1, \dots, b_m]$ is the bias value of hidden neuron and  is the desired target in the training set. In training set, ELM with n neurons in hidden layer approximates input samples with zero error such that $\sum_{j=1}^{m}||t'_j-{t}_j||=0$ where ${t'}_j$ is network output computed with using $w'$ in (\ref{elm3}). But in this case due to over-fitting generalization capacity becomes very poor.

\label{ELM1}

Hidden layer neuron number m must be selected randomly or empirically, such that m<n to prevent overfitting. Inverse of $\mathbf{H}$ can not be determined directly if  $\mathbf{H}$ is not a full-rank matrix. Pseudoinverse of the matrix $\mathbf{H}$, namely $\mathbf{H}^+$, can be computed via least square solution. To stabilize the pseudoinverse numerically, regularized least squares solution in (\ref{elm4}) is used,

\begin{equation}
    \label{elm4}
    \mathbf{H}^{+} = \left\{
        \begin{array}{ll}
             (\mathbf{H}^{T}\mathbf{H} + \lambda  I )^{-1} \mathbf{H}^T. & \quad L \leq N \\
           (\mathbf{H}\mathbf{H}^T+ \lambda I)^{-1}\mathbf{H}^T & \quad L > N
        \end{array}
\right\}
\end{equation}
where $\lambda$ is regularization parameter that enables linear independence of columns of $\mathbf{H}^T\mathbf{H}$. This solution is accurate as long as square matrix $(\mathbf{H}^{T}\mathbf{H} + \lambda I)$ is invertible. In SLFN, it is singular in most of the cases since there is tendency to select m<<n. In conventional ELM, Huang et. al. \cite{ref22} has solved this problem using SVD \cite{ref23} method. However, SVD is very slow when dealing with large data and has low-convergence to real solution \cite{ref24,ref25}.

\subsection{Hessenberg Decomposition ELM $-$HessELM}
Since (\ref{elm3})  is an under-determined system of equation,  pseudoinverse of hidden layer output matrix $\mathbf{H}$ is formed as \footnote{Hessenberg decomposition works only for square matrices, therefore least square solution is needed.}
\begin{equation}
    \mathbf{H}^{+}=(\mathbf{H}\mathbf{H}^T)^{-1}\mathbf{H}^T
\label{hesselm1}
\end{equation}
to reach $\mathbf{H}^{+}$, square matrix $\mathbf{H}\mathbf{H}^T$ can be decomposed using hessenberg decomposition.
\begin{equation}
    \mathbf{H}\mathbf{H}^T=QUQ^{*}
    \label{hesselm2}
\end{equation}
where Q  is a unitary matrix and U is an upper hessenberg matrix. When $\mathbf{H}\mathbf{H}^T$ is substituted in (\ref{hesselm1})

\begin{align}
\label{hesselm3}
\begin{split}
            \mathbf{H}^{+} &=(QUQ^{*})^{-1} \mathbf{H}^T\\
             &=QU^{-1}Q^{*} \mathbf{H}^T
\end{split}
\end{align}

where U is an upper hessenberg matrix that is also symmetric and tridiagonal. When considering singularity conditions, it is reached that $\begin{vmatrix} U \end{vmatrix} \neq 0$ and matrix U is non-singular therefore, $U^{-1}$ exists. Target values are reached and learning are achieved as follows, output weights are reached and put in its place in \ref{elm3}.

\subsection{Leave-One-Out (LOO) Error Based Optimization}
Leave-one-out (LOO) is a parameter optimization and model selection method used in machine learning. LOO method is used to select optimum regularization parameter $\lambda$ in (\ref{elm4}) that minimizes mean square error predicted residual sum of squares $MSE^{PRESS}$ in \ref{loo1}. $MSE^{PRESS}$ is calculated as,

\begin{equation}
    MSE^{PRESS}=\frac{1}{N}\sum_{j=1}^N(\frac{t'_j-t_j}{1-HAT_{jj}})^2
    \label{loo1}
\end{equation}
where $HAT_{jj}$ is  diagonal $j^{th}$ value on diagonal of $HH^+$. Using \ref{hesselm1} HAT is calculated as,

\begin{equation}
    HAT = \left\{
        \begin{array}{ll}
            \mathbf{H}(\mathbf{H}^T\mathbf{H} + \lambda I )^{-1}\mathbf{H}^T & \quad L \leq N \\
            \mathbf{H}^T(\mathbf{H}\mathbf{H}^T+ \lambda I)^{-1}\mathbf{H}^T  & \quad L > N
        \end{array}
\label{loo2}
\right\} 
\end{equation}
Cao et. al \cite{ref18} introduced implementation of $MSE^{PRESS}$ in ELM with SVD method in (\ref{loo3}).

\begin{align}
     \label{loo3}
     \begin{split}
            HAT=\mathbf{H}V(D^2 + \lambda I)^{-1}V^T\mathbf{H}^T  
     \end{split}
\end{align}

H matrix is decomposed into  using SVD method. Note that in \ref{loo3} calculation of HAT is irrelevant with U. In this study implementation of  in ELM with hessenberg decomposition method was introduced MATLAB scripts for regularized hessenberg decomposition based ELM (R-HessELM) is shared for practitioners \footnote{https://github.com/apdullahyayik/Regularized-HessELM}

\begin{equation}
    \label{loo4}
    \begin{split}
    HAT &=H^T(\mathbf{H}\mathbf{H}^T+ \lambda I)^{-1}\mathbf{H}^T \\
    &=\mathbf{H}^T(QUQ^* +  \lambda I)^{-1}\mathbf{H}^T \\
    &= \mathbf{H}^T Q(U+\lambda I)^{-1} Q^T
    \end{split}
\end{equation}

\begin{table*}
\centering
\caption{Performances of CHF prediction}
\label{Table1}
\begin{tabular}{@{}llccc@{}}
\toprule
Feature Model                     & Classifier   & Precision (\%) & Sensitivity (\%) & Overall Accuracy (\%) \\ \midrule
\multirow{4}{*}{Circled Entropy}  & ELM          & 92.05          & 93.25            & 92.42                  \\
                                  & R-ELM ($\lambda=e^{-10}$)     & 93.37          & 94.36            & 93.43                  \\
                                  & HessELM      & 92.73          & 93.91            & 93.93                  \\
                                  & R-HessELM ($\lambda=e^{-11}$) & 97.26          & 97.77            & 96.36                  \\
\multirow{4}{*}{Inclined Entropy} & ELM          & 97.2           & 97.87            & 97.98                  \\
                                  & R-ELM ($\lambda=e^{-10}$)     & 97.09          & 98.04            & 98.48                  \\
                                  & HessELM      & 96.95          & 97.39            & 97.87                  \\
                                  & R-HessELM ($\lambda=e^{-12}$) & 98.05          & 98.3             & \textit{98.49}                  \\
\multirow{4}{*}{Squared Entropy}  & ELM          & 95.37          & 96.06            & 95.45                  \\
                                  & R-ELM ($\lambda=e^{-13}$)     & 96.69          & 97.12            & 95.55                  \\
                                  & HessELM      & 96.32          & 97.09            & 96.26                  \\
                                  & R-HessELM ($\lambda=e^{-16}$) & 97.47          & 97.98            & 96.76                  \\
\multirow{4}{*}{Grid Entropy}     & ELM          & 91.32          & 92.85            & 95.05                  \\
                                  & R-ELM ($\lambda=e^{-13}$)     & 92.41          & 93.73            & 95.96                  \\
                                  & HessELM      & 92.87          & 94.31            & 95.75                  \\
                                  & R-HessELM ($\lambda=e^{-18}$) & 93.98          & 95.15            & 96.16                  \\ \cmidrule(l){1-5} 
\end{tabular}
\end{table*}

Algorithm 1 provides the detail implementation of LOO based output layer weight matrix calculation with regularized HessELM.

\begin{algorithm}
\caption{Computing weights with regularized HessELM}\label{alg:euclid}
\begin{algorithmic}[1]\\
Calculate Hessenberg Decomposition of $\mathbf{H}\mathbf{H}^T$, $\mathbf{H}\mathbf{H}^T=QUQ^*$
\Procedure{Compute Weights}{$H, Q, U, t, \lambda_{candid\_set}$}
\While{$\lambda_i \in \lambda_{candid\_set}$} 
\State $t'=\mathbf{H}(Q(U+\lambda_i I)^{-1}Q^TH^T t)$
\State $R=t'-t$
\State $HAT= \mathbf{H}^T Q(U+\lambda I)^{-1} Q^T$
\State $S=\textbf{diag}(\textbf{eye}(size(HAT,2))-HAT)$
\State $MSE^{PRESS}_i=(R./S)^2$
\EndWhile{\textbf{end}}
\State \textbf{find} $\lambda_{opt}$ corresponding to \textbf{min}($MSE^{PRESS}$)
\State $w=Q(U+\lambda_{opt} I)^{-1}Q^TH^T t$\\
\Return $w$
\EndProcedure
\end{algorithmic}
\end{algorithm}

\begin{figure*}
  \subfigure[]{%
    \label{fig:A1}%
    \includegraphics[width=0.33\textwidth]{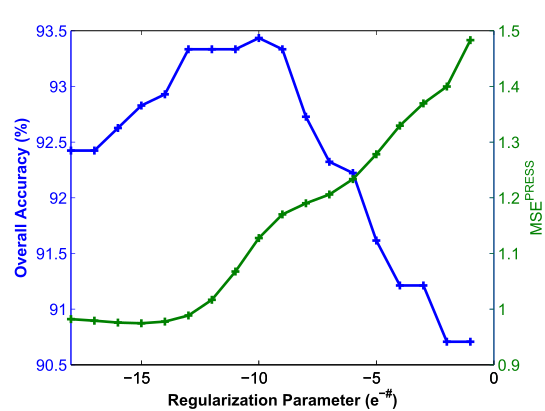}
            }%
  \subfigure[]{%
    \label{fig:B1}%
   \includegraphics[width=0.35\textwidth]{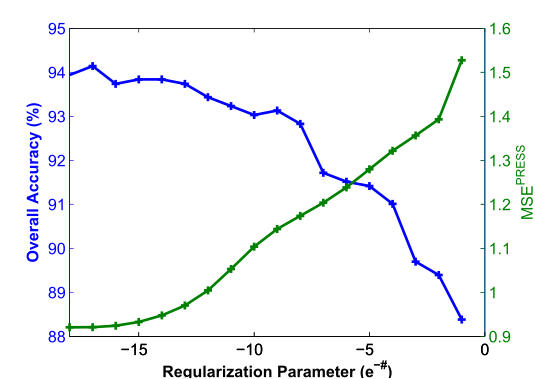}
            }\\
  \subfigure[]{%
    \label{fig:C1}%
    \includegraphics[width=0.35\textwidth]{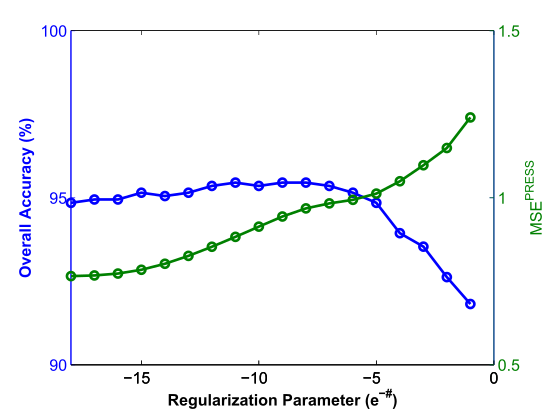}
            }%
  \subfigure[]{%
    \label{fig:D1}%
    \includegraphics[width=0.35\textwidth]{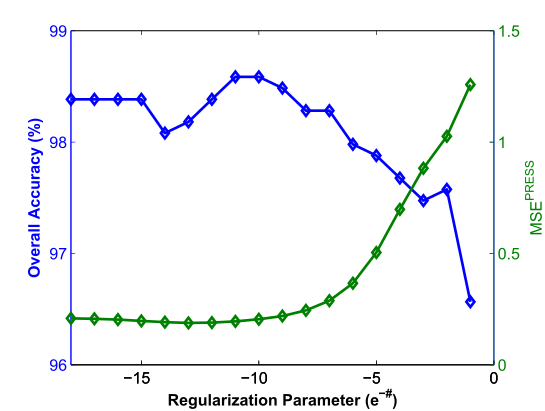}
           }\\
  \subfigure[]{%
    \label{fig:E1}%
    \includegraphics[width=0.35\textwidth]{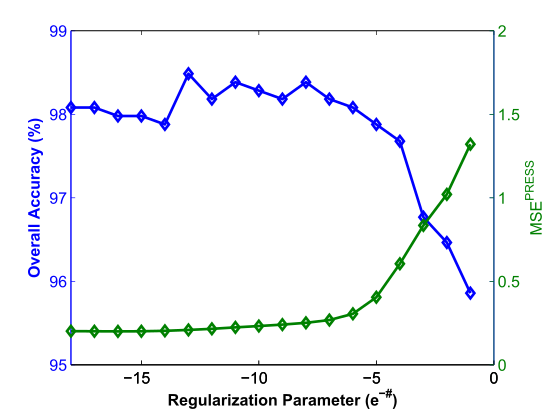}
            }%
  \subfigure[]{%
    \label{fig:F1}%
    \includegraphics[width=0.35\textwidth]{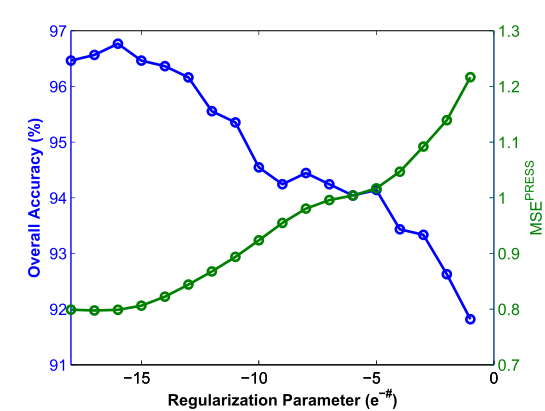}
            }\\
  \subfigure[]{%
    \label{fig:G1}%
    \includegraphics[width=0.35\textwidth]{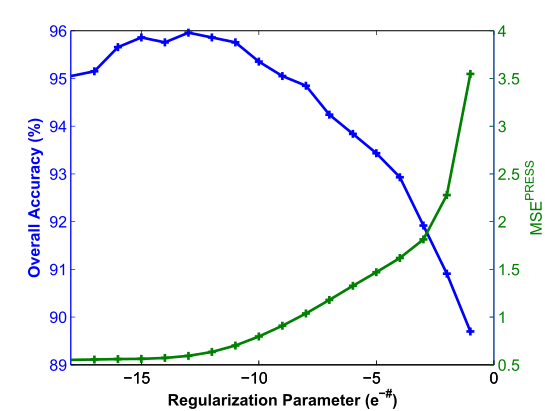}
            }%
  \subfigure[]{%
    \label{fig:H1}%
    \includegraphics[width=0.35\textwidth]{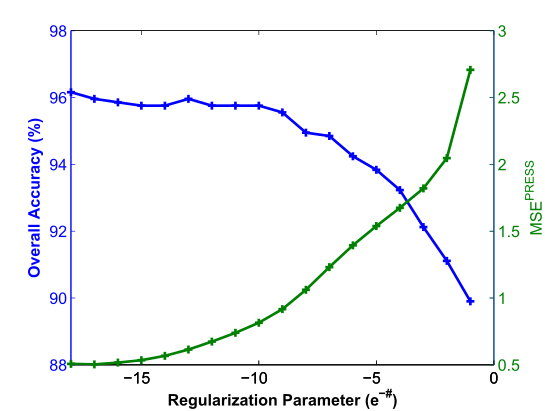}
            }
  \caption{Overall accuracy and $MSE^{PRESS}$ values with regard to regularization parameter of ELM with (a.1)  circled, (b.1) squared, (c.1) inclined (d.1) grid entropy measurements and HessELM with (a.2)  circled, (b.2) squared, (c.2) inclined (d.2) grid entropy measurements}%
  \label{figure5}%
\end{figure*}

\section{Results and Conclusions}
\label{resultsandconclusions}
In this study we proposed pattern recognition approach for predicting CHF medical condition from ECG recordings. Optimally selection of regularization parameter of ELM with SVD in \cite{ref18} was extended to ELM with hessenberg decomposition introduced in \cite{ref19}. The designed system used 4 number of entropy measurements, CEM, SEM, IEM and GEM of SODP in ECG time-series, as discriminative features. The presented R-HessELM, which uses IEM features resulted in an overall accuracy of 98.49\%.

In Figure \ref{figure5}, the effect of regularization parameter (between $e^{-20}$ and $e^{-1}$) on both overall accuracy and $MSE^{PRESS}$ value for ELM and HessELM classifiers with proposed feature models are shown. One can see that larger regularization parameter yields to  $MSE^{PRESS}$ larger  value and lower performance accuracy. As it is suggested  $MSE^{PRESS}$ and performances are inversely proportional with each other \ref{result1} in particular between $e^{-15}$ and $e^{-1}$. If regularization parameter converges to zero, performances do not rise although $MSE^{PRESS}$ value decreases. Therefore, in proposed classifier model, first task should be selecting optimum regularization parameter with regard to only minimum $MSE^{PRESS}$  value, not performances to reduce complexity.

\begin{equation}
    \label{result1}
    MSE^{PRESS} \propto Performances
\end{equation}

\begin{table*}
\centering
\caption{Comparisons of studies}
\label{table2}
\begin{tabular}{@{}llllll@{}}
\toprule
Study                   & Class                                                                                             & Feature Model                                                                                                                 & Feature Size & Classifier                                                             & Overall Accuracy (\%)                                            \\ \midrule
Asyal{\i}, 2003  \cite{ref35}          & Normal CHF                                                                                        & \begin{tabular}[c]{@{}l@{}}Standard Deviation\\ Normal to Normal\end{tabular}                                                 & -            & BayesNET                                                               & 89.95                                                            \\
\.{I}\c{s}ler and Kuntalp, 2007 \cite{ref13}  & Normal CHF                                                                                        & \begin{tabular}[c]{@{}l@{}}Wavelet Entropy\\ Poincare Plot\\ Fast Fourier  Transform\\ Genetic feature selection\end{tabular} & 9            & k-nn                                                                   & 93.98                                                            \\
Ubeyli, 2009   \cite{ref36}         & \begin{tabular}[c]{@{}l@{}}NormalCHF\\ Ventricular Arrhythmia \\ Atrial Fibrillation\end{tabular} & Eigenvector Method                                                                                                            & -            & \begin{tabular}[c]{@{}l@{}}Recurrent Neural\\ Network\end{tabular}     & 98.06                                                            \\
Thuraisingham, 2009 \cite{ref33}    & Normal CHF                                                                                        & \begin{tabular}[c]{@{}l@{}}SODP Central \\ Tendency\end{tabular}                                                              & -            & k-nn                                                                   & Almost 100\%                                                     \\
Kamath, 2012 \cite{ref34}            & Normal CHF                                                                                        & \begin{tabular}[c]{@{}l@{}}Teager Energy of \\ Poincare Plot\end{tabular}                                                     & -            &                                                                        & Almost 100\%                                                     \\
Yu and Lee, 2012 \cite{ref26}       & Normal CHF                                                                                        & \begin{tabular}[c]{@{}l@{}}Bispectrum-related \\ features, Genetic \\ feature selection\end{tabular}                          & -            & SVM                                                                    & 96.38                                                            \\
Yu and Lee, 2012 \cite{ref27}        & Normal CHF                                                                                        & \begin{tabular}[c]{@{}l@{}}UCIMFS, MIFS,CMIFS,\\ mRMR and\\ MI-based greedy\\ feature selection\end{tabular}                  & 15           & SVM                                                                    & 97.59                                                            \\
Orhan, 2013  \cite{ref32}           & Normal CHF                                                                                        & \begin{tabular}[c]{@{}l@{}}Equal Frequency in \\ Amplitude(EFiA)\\ Equal Width in Time\\ (EWiE)\end{tabular}                  & -            & \begin{tabular}[c]{@{}l@{}}Linear \\ Regression\end{tabular}           & 99.3                                                             \\
Liu et al., 2014 \cite{ref28}      & Normal CHF                                                                                        & \begin{tabular}[c]{@{}l@{}}Three nonstandard\\ HRVmeasures\\ (i.e. SUM\_TD, \\ SUM\_FD and \\ SUM\_IE)\end{tabular}           & -            & SVM                                                                    & 100                                                              \\
Narin et. al, 2014 \cite{ref29}     & Normal CHF                                                                                        & \begin{tabular}[c]{@{}l@{}}Wavelet Transform\\ Backward elimination\\ Method\end{tabular}                                     & -            & SVM                                                                    & 91.56                                                            \\
Heinze et al., 2014 \cite{ref30}    & Normal CHF                                                                                        & Power spectral Density                                                                                                         & -            & \begin{tabular}[c]{@{}l@{}}Learning Vector\\ Quantization\end{tabular} & \begin{tabular}[c]{@{}l@{}}13.6\% error\\ at 50 min\end{tabular} \\
Majahan et al., 2017 \cite{ref31}   & Normal CHF                                                                                        & \begin{tabular}[c]{@{}l@{}}Probabilistic symbol\\ pattern Recognition\end{tabular}                                            & -            & \begin{tabular}[c]{@{}l@{}}Decision\\ Trees\end{tabular}               & 99.5                                                             \\
Acharya et al., 2017 \cite{ref37}      & Normal CHF                                                                                        & \begin{tabular}[c]{@{}l@{}}Empirical mode \\ Decomposition\end{tabular}                                                       & 13           & SVM                                                                    & 97.64                                                            \\
This study              & Normal CHF                                                                                        & \begin{tabular}[c]{@{}l@{}}Incline Entropy  \\ Measures (IEM) \\ of SODP\end{tabular}                                         & 17           & \begin{tabular}[c]{@{}l@{}}Regularized\\ HessELM\end{tabular}          & 98,41                                                            \\ \bottomrule
\end{tabular}
\end{table*}

5 fold cross-validated classification performances with selected specific regularization parameter are detailed in Table \ref{Table1}. The highest overall accuracy was achieved with 98.49\%, precision with 98.05\% and sensitivity with 98.30\% using hessenberg decomposition based extreme learning machine ($\lambda =e^{-16}$) and IEM features. Since amplitude and interval of R peaks have significant and discriminant role in CHF medical condition of ECG recordings, IEM features that capture wide-range of changes, in particular R peaks,  result in better performance than other approaches for CHF prediction.

\section{Discussion}
\label{discussion}
Our study shows that the use of regularized HessELM which considers effective pseudoinversing to predict the medical condition of ECG signal can yield considerably high performances. Entropy measurements of SODP on Cartesian system aimed at describing the amplitude  distributions of ECG time-series.

As suggested by the SODPs in Figure \ref{figure3} it appears to be a difference on data distribution, with normal recordings fewer points at the center of Cartesian system than CHF ones. This tendency is captured by the utilized entropy measurements, which works as effective predictors. The early predictions of  CHF result in successfully managing the treatment process and preventive revulsion for medical purposes. The predisposition of the embedded systems into medical processes with high prediction performance rates of precision, sensitivity, and overall accuracy provides enhancing the diagnostic tools. Consistency of performance observed would imply the possibility of designing embedded systems of CHF predictors.

HessELM model has the ability to perform more accurate performances than  conventional one using various entropy measurements on SODP. Furthermore, regularized approaches of both models could achieve high performance  when regularization parameter was chosen between e-10 and e-18. The SODP dispersion of the ECG signals from both healthy and CHF subjects pictures as an elliptical-shaped form on Cartesian system. IEM features enable most detailed quantization of the SODP considering the characteristics of such elliptical-shape. In other words, it performs segmentation of SODP depending on the slope of the propagation. That is why it is the most responsible entropy measurement for CHF predictions on ECG signals.
To improve the performance rate of extreme learning machines, further investigations need to be performed. For instance, a robust feature selection algorithm that provides rejection of redundant information from ECG data, may lead to even higher prediction performances.

\section*{\small{Compliance with ethical standards}}
\textbf{\small{Conflict of interest}} \small{The authors declare that there is no conflict of interest.}

\bibliographystyle{spmpsci}      

%
%

\end{document}